\pdfoutput=1

\documentclass[11pt]{article}

\usepackage[]{ACL2023}

\usepackage{times}
\usepackage{latexsym}

\usepackage[T1]{fontenc}

\usepackage[utf8]{inputenc}

\usepackage{microtype}

\usepackage{inconsolata}

\usepackage{footmisc}
\usepackage{graphicx}
\usepackage{multirow}
\usepackage{tabularx}
\usepackage{csquotes}
\usepackage{booktabs}
\usepackage{bm}
\usepackage{CJKutf8}
\usepackage{array}
\usepackage{latexsym}
\usepackage{amsthm,amsmath,amssymb}
\usepackage{array}
\usepackage{url}
\usepackage{float}

\usepackage{xcolor}
\usepackage{enumitem}
\usepackage[normalem]{ulem}

\def\ODdel#1{\bgroup\markoverwith{\textcolor{cyan!89!yellow!80!black!100}{\rule[0.4ex]{2pt}{3pt}}}\ULon{#1}}

\title{Text Style Transfer: An Introductory Overview}

\author{Sourabrata Mukherjee \and Ondřej Dušek \\
  Charles University, Faculty of Mathematics and Physics \\
  Institute of Formal and Applied Linguistics \\
  Prague, Czech Republic \\
  \texttt{\{mukherjee,odusek\}@ufal.mff.cuni.cz} \\}

\begin{document}
\maketitle
\begin{abstract}
Text Style Transfer (TST) is a pivotal task in natural language generation to manipulate text style attributes while preserving style-independent content. The attributes targeted in TST can vary widely, including politeness, authorship, mitigation of offensive language, modification of feelings, and adjustment of text formality. TST has become a widely researched topic with substantial advancements in recent years. This paper provides an introductory overview of TST, addressing its challenges, existing approaches, datasets, evaluation measures, subtasks, and applications. This fundamental overview improves understanding of the background and fundamentals of text style transfer.
\end{abstract}

\section{Introduction}
\paragraph{Natural Language Generation}
Natural Language Generation (NLG) is the process of producing meaningful phrases and sentences in natural language.
The main goal of NLG is to automatically produce narratives that describe, summarize, and explain the input data in a human-like manner. In other words, it generates fluent texts with minimal grammatical errors and retains the specific intended content.



Some of the popular NLG tasks include machine translation \cite{cho2014learning}, dialogue systems \cite{shang2015neural}, and text summarization \cite{rush2017neural}. Through these tasks, the generated text has shown to be more coherent, logical, and emotionally rich, 
especially with the latest approaches based on neural language models.


\paragraph{Controllable NLG}

Most of the built NLG systems target text fluency and grammatical correctness, and do not consider any specific control over text style. This is a motivation for research on controllable text generation 
\cite{len2020controllable}. 
The aspects of text generation that are commonly controlled include topic \cite{dziri-etal-2019-augmenting,feng2018topic,ijcai2018-619,xing2017topic}, style \cite{li2018delete, sudhakar2019transforming, prabhumoye2018style, chen2018adversarial}, emotion \cite{fu2018style,kong2019adversarial,DBLP:journals/inffus/SunLWLT20,zhou2018emotional}, and user preferences \cite{li-etal-2016-persona,luan-etal-2017-multi,yang2018investigating,yang2017personalized}. Some of the applications of controllable text generation are context-based text generation \cite{jaech2018low}, topic-aware text generation, \cite{wang2018reinforced}, knowledge-enhanced text generation \cite{young2018augmenting} and text style transfer \cite{hu2022text}.

Control can be applied at various stages of the neural generation process, such as input, hidden states, and decoding \cite{prabhumoye2020exploring}. The Plug and Play language model (PPLM) that was proposed by \citet{dathathri2019plug} takes an external input, performs computations on hidden states, and then combines a pre-trained language model with one or more simple attribute classifiers that guide text generation toward the desired topic or sentiment. Another model by \citet{gehrmann2018end} describes a training method based on diverse ensembling that would lead models to learn distinct text styles. It can thus be inferred that end-to-end models can be equipped with the ability to control style and length. More details on how NLG can be controlled using various control strategies in the state-of-the-art models can be found in \cite{len2020controllable}.


\paragraph{Style-Controlled Text Generation}

In recent research, more attention has been paid to a subtask of controllable text generation dubbed \textit{style-controlled text generation}, i.e., modeling and manipulating the style of the generated text \cite{mou2020stylized}. The goal of this approach is to model the content of a text along with controlling its style. For example, the persona of a speaker in dialogue \cite{li2016persona} or the sentiment of product reviews \cite{hu2017toward}. Understanding and dealing with style in text proves to be very complex \cite{hu2022text}, but recent advances in deep learning techniques are helping stylized text generation tasks in various ways \cite{jin2022deep}. For example, embedding learning techniques are used to represent style \cite{fu2018style}, and then adversarial learning is used to match content but to distinguish between different styles\cite{hu2017toward, xu2018unpaired, john2018disentangled}.

\paragraph{Text Style Transfer}

In this paper, we will focus on Text Style Transfer (TST). TST is a task closely related to Style-Controlled Text Generation. \textit{Style-Controlled Text Generation} aims to generate new text in a specific style. In contrast, \textit{Text Style Transfer} is an existing text written in source style, aiming to change the text style, i.e. a text retaining most of the content but conforming to the target style.
Our aim is to give a very basic introduction to the TST task. All of the sections are presented in a brief and simple manner with an illustrative number of examples. A more detailed overview can be found in \cite{hu2022text, jin2022deep, mou2020stylized, toshevska2021review, prabhumoye2020exploring}.

The paper is organized as follows. After the introductory section, Section \ref{tst} provides an overview of text style transfer. 
Section~\ref{challenges} reflects on the challenges facing the TST task. The discussion of the existing data sets, approaches, evaluation measures and applications is presented in Sections~\ref{datasets}-\ref{applications}. A short overview of the related ethical considerations is given in Section \ref{ethics}. Section \ref{conclusion} concludes the paper.


\section{The Task} \label{tst}

Text style transfer (TST) is an NLG task that aims to automatically control the style attributes of a text while preserving the style-independent content. Some of the attributes that TST aims to control are politeness, formality, sentiment, and many others. Table \ref{tbl:TSTexamples} shows some basic examples of TST. 
TST implies the need to understand the difference between the style and content of a text.

\begin{table*}[t]
\begin{tabular}{lll}
\toprule
& {\textbf{Source} \color{red}\textbf{Style}} & {\textbf{Target} \color{blue}\textbf{Style}} \\
\midrule

Impolite \textrightarrow{} Polite$\colon$ & \textcolor{red}{Shut up!} the video is starting! & \textcolor{blue}{Please be quiet}, the video will begin shortly. \\

Negative \textrightarrow{} Positive$\colon$ & The food is \textcolor{red}{tasteless}.  & The food is \textcolor{blue}{delicious}. \\

Informal \textrightarrow{} Formal$\colon$ & The kid is  \textcolor{red}{freaking out}. & That child is \textcolor{blue}{distressed}.\\

\bottomrule
\end{tabular}
\caption{TST examples regarding sentiment, polarity, and formality.}
\label{tbl:TSTexamples}
\end{table*}


\subsection{Understanding Style and Content} \label{styleandcontent}

\citet{mcdonald1985computational} define style as a notion that refers to the manner in which semantics is expressed. Individualistic styles such as choice of words, sentence structures, metaphors, sentence arrangement, etc., vary from person to person. These variations are shaped by the speakers's personality -- everyone has a distinctive set of techniques for using the language to express and achieve their independent goals \cite{hovy1987generating}. This individualistic nature also determines how a person perceives events, describes ideas, or provides additional information about them \cite{hu2022text}. 
Style extends beyond individual sentences to the broader discourse level. This includes elements such as paragraph organization, theme progression, and use of cohesive devices that bind the text together. These stylistic features at the discourse level play a crucial role in ensuring that the text is coherent and engaging, thereby enhancing its ability to convey the intended message and intrigue the reader. Taking these aspects into account, the text can offer a richer and more nuanced understanding of its content.

Style has also been defined by \citet{hovy1987generating} by its pragmatic aspects. 
Beyond these personal styles of expression, there are certain styles that are used as protocols to regularize the manner of communication. For example, in the case of academic writing, using formal expressions is the regularized protocol. 

TST studies adopt a more data-driven approach to define text style in contrast to the theoretical definition used in linguistic studies \cite{jin2022deep}. 
We can define style in TST as the text style attributes or labels that are dependent on style-specific corpora \cite{hu2022text}. For example, datasets are  manually annotated with linguistic style definitions, such as formality \cite{rao2018dear} or sentiment \cite{shen2017style,he2016ups,dai2019style}. Unfortunately, not all possible styles have very well-matched corpora, and many recent dataset collection works are looking for meta-information that would automatically link a corpus to a certain style.
Some of the TST tasks are built upon the assumption that style is localized to certain tokens in a text, and a token has either content or style information, but not both \cite{lee2021enhancing}.

In opposition to style, content can be understood as the subject matter, theme, or topics the author writes about. 

\subsection{Problem Formulation} \label{problemformulation}
\label{sec:problem-formulation}

Given a text $x$, with an original style $S$, our goal is to rephrase $x$ into a new text $\hat{x}$ with a target style $S'$ ($S' \neq S$) while preserving its content that is independent of style.

Suppose that we have a dataset $X_S = { x_{1}^{(S)}, \dots, x_{m}^{(S)} }$ representing texts in style $S$. The task is to transform texts in style $S$ to the target style $S'$ while maintaining the original meaning. We denote the output of this transformation by $X_{S \rightarrow S'} = { \hat{x}{1}^{(S')}, \dots, \hat{x}{m}^{(S')} }$. Similarly, for the inverse transformation from style $S'$ to style $S$, we denote the output as $X_{S' \rightarrow S} = { \hat{x}{1}^{(S)}, \dots, \hat{x}{n}^{(S)} }$.

\section{Challenges} \label{challenges}

Modeling the style of text comes with a lot of challenges in practice, which are discussed in this section.
\paragraph{No Parallel Data} \label{noparalleldata}

TST models could be trained with respect to parallel text from a given style or on non-parallel corpora. Parallel datasets are those which consist of pairs of texts (i.e. sentences, paragraphs) where each text in the pair expresses the same meaning, but in a different style. Non-parallel datasets, on the other hand, have no paired examples to learn from, and simply exist as mono-style corpora. For parallel datasets, TST can be formulated in such a way that instead of translating between languages, one can translate between styles following machine translation. However, obtaining suitable, sufficient parallel data for each desired style attribute is the biggest challenge. 

\paragraph{Style and content are hard to separate} \label{hardstylecontent}

Style transfer text generation implies the need to distinguish content from style. In some scenarios, the line between content and style can be blurry.
This is since the subject on which an author is writing can also influence their choice of words and style. This interweaving of the style and semantics makes TST challenging.   

\paragraph{No Standard Evaluation Measures}

Evaluating the quality of the style-transferred text is hard. Human evaluation is regarded as the best indicator of quality, but unfortunately, it is expensive, slow, and hard to reproduce \cite{belz2020reprogen}, making it an infeasible approach to use on a daily basis to validate model performance. For this reason, we often rely on automated evaluation metrics to serve as a cheap and quick proxy for human judgment.

In the case of automatic evaluation of TST, it has been noticed that when style transfer accuracy increases, the content preservation scores decrease, and vice versa \cite{mukherjee2022balancing}. The main reason behind this is the entanglement between the content and style (see above). This trade-off between style transfer accuracy and content preservation poses a very big challenge for evaluating TST tasks. 

In order to effectively evaluate a TST output, one must pay attention to how semantically accurate the output text is and how fluent it is. The comprehensive TST evaluation also considers three criteria: transferred style accuracy, semantic preservation, and fluency, which often require human evaluation as automated metrics alone do not adequately identify these complex properties. Further discussion on evaluation measures is in Section~\ref{eval}.

\section{Datasets and Benchmarks} \label{datasets}

To evaluate TST models, many datasets have been proposed over the years. We discuss a few popular datasets by individual subtasks as follows:

\paragraph{Politeness Transfer} 
Politeness transfer aims to control the politeness of a text \cite{mukherjee-etal-2023-polite, madaan2020politeness}. A compiled dataset with automatically labeled instances from the raw Enron e-mail corpus \cite{shetty2004enron} was presented by \citet{madaan2020politeness}. This dataset mainly focuses on politeness in North American English.

\paragraph{Sentiment Transfer}
Another common task in TST is sentiment transfer (transferring text's polarity from positive to negative or vice-versa) \cite{mukherjee-etal-2023-low, mukherjee2022balancing}. There are three popular datasets proposed for this task.

\begin{itemize}
\item  Yelp \textendash \ This is a corpus consisting of restaurant reviews from Yelp collected by \citet{shen2017style}. 
\item Amazon \textendash \ This is Amazon's product reviews that were collected by \citet{he2016ups}. 
\item IMDb \textendash \ This is a movie review dataset constructed by \citet{dai2019style}.
\end{itemize}

\paragraph{Formality Transfer}
Formality transfer is yet another task in TST which is not only complex but also involves multiple attributes that affect text formality. Grammarly’s Yahoo Answers Formality Corpus (GYAFC) is the largest human-labeled parallel dataset that was proposed for formality transfer tasks by \citet{rao2018dear}. The authors extracted informal sentences from the Entertainment \& Music and Family \& Relationship domains of the Yahoo Answers L6 corpus for preparing the dataset. 

\paragraph{Author's Style Re-writing}
The task of paraphrasing a sentence to match a specific author's style is called author imitation. To tackle such tasks, \citet{xu2012paraphrasing} collected a parallel dataset that captured line-by-line modern interpretations of 16 Shakespeare's plays, with the help of the educational site Sparknotes.\footnote{\url{https://www.sparknotes.com}} The objective behind collecting the dataset was to imitate Shakespeare’s text style by transferring modern English sentences into Shakespearean-style sentences. This dataset has been used in other TST studies as well \cite{jhamtani2017shakespearizing,He2020A}.


\paragraph{Image Captions Transfer}
The task of transferring image captions from factual formal ones to romantic and humorous styles was proposed by \citet{li2018delete}. Following this, a caption dataset was collected by the authors where each sentence was labeled as factual, romantic, or humorous.

\paragraph{Text Simplification}
Another important use of TST is to lower the language barrier for readers, which includes tasks like converting general English into Simple English, based on a dataset collected from Wikipedia \cite{zhu2010monolingual}. Another task is to simplify medical descriptions to patient-friendly text \cite{van2019evaluating}. 


\paragraph{Political-slant Transfer}
Political slant transfer is a task that modifies a writer's political affiliation writing style while preserving the content. Comments from Facebook posts from 412 members of the United States Senate and House who have public Facebook pages were collected by \citet{prabhumoye2018style} and further annotated with each congressperson’s political party affiliation, i.e., Democrat or Republican. 

\paragraph{Fixing offensive texts}
Correcting offensive and abusive language \cite{sourabrata-etal-2023-text} is another important task of TST which is a major problem in today's world, due to the prevalence of abusive comments on social media. Posts from Twitter and Reddit were collected by \citet{dos2018fighting} and then classified into \emph{offensive} and \emph{non-offensive} classes using a classifier pre-trained on an annotated offensive language dataset.



\section{Text Style Transfer Approaches} \label{approaches}

Standard data-driven TST approaches can be classified based on the data used for training (parallel vs.\ non-parallel). Recently, new approaches using large language models (LLMs) emerged that do not specifically need in-domain training data.

\subsection{Supervised Training on Parallel Data}

For situations where style-parallel data is available, like most supervised methods, a standard sequence-to-sequence model \cite{sutskever2014sequence, mukherjee-etal-2023-low, mukherjee2024multilingual} with the encoder-decoder structure is typically used 
\cite{hu2022text}. This process is similar to machine translation and text summarization. The encoder-decoder architecture can be implemented by either LSTM \cite{hochreiter1997long} or the Transformer \cite{vaswani2017attention} architecture. For example, \cite{jhamtani2017shakespearizing} trained a sequence-to-sequence model on a parallel corpus and then applied the model to translate modern English phrases to Shakespearean English. However, the application of basic sequence-to-sequence approaches is quite limited due to the lack of parallel data (see Section \ref{noparalleldata}).

\subsection{Non-parallel Approaches}

Methods applicable to non-parallel data can broadly be divided into three unsupervised approaches:

\paragraph{Prototype Editing} This process works by deleting only the parts of the sentences which represent the source style and replacing them with words with the target style while making sure that the resulting text is still fluent. The advantage of this approach is its simplicity and explainability. For example, \cite{li2018delete, mukherjee2023low} found that parts of a text that are associated with the original style can be replaced with new phrases associated with the target style. The text was then fed into a sequence-to-sequence model to generate a fluent text sequence in the target style. However, these approaches are not suitable for TST applications where simple phrase replacement is not enough or a correct way to transfer style. The style marker retrieval might not work if the datasets have confounded style and contents. This is because they may lead to the incorrect extraction of style markers, affecting some content words. 

\paragraph{Disentanglement} This approach aims at disentangling the text into its content and style in an embedding latent space, then applies generative modeling.
TST models first learn the latent representations of the content and style of the given text. The latent representation of the original content is then combined with the latent representation of the desired target style to generate text in the target style. Techniques such as back-translation \cite{mukherjee2022balancing,zhang2019machine,prabhumoye2018style} and adversarial learning \cite{shen2017style,zhao2018adversarially,fu2018style} have been proposed to disentangle latent representations into content and style.
In general, total disentanglement is impossible without inductive biases or some other forms of supervision \cite{locatello2019challenging}. 

\paragraph{Pseudo-Parallel Corpus Creation} This process is used to train the model in a supervised way by generating pseudo-parallel data. One way of constructing pseudo-parallel data is through retrieval, i.e., extracting aligned sentence pairs from two mono-style corpora. \citet{jin2019imat} constructed pseudo-parallel corpora by matching sentence pairs in two style-specific corpora according to cosine similarity over pre-trained sentence embeddings. 
The constructed pseudo-parallel corpora must reach a certain level of quality to be useful for TST.

\subsection{Using Large Language Models}
\label{sec:llms}

LLMs have revolutionized the field of natural language processing by generating coherent and contextually relevant text \cite[e.g.,][]{touvron2023llama,touvronLlamaOpenFoundation2023}. By learning from vast amounts of text data, LLMs capture various linguistic styles and nuances. This capability is particularly beneficial for TST tasks.

A distinctive feature of LLMs is their ability to perform valuable tasks without fine-tuning, showcasing zero- and few-shot capabilities \cite{Liu:2023}. Style transfer has been framed as a sentence rewriting task, enhancing LLMs' zero-shot performance for arbitrary TST by using task-related exemplars \cite{reif2021recipe}. A reranking method has been proposed to select high-quality outputs from multiple candidates generated by the LLM, thereby improving performance \cite{suzgun2022prompt}. Additionally, dynamic prompt generation has been introduced to guide the language model in producing text in the desired style \cite{liu2024adaptive}.

While prompt engineering is the prevalent approach \cite{brown2020language, jiang2020can}, LLMs are highly sensitive to prompts \cite{mishra2021reframing, zhu2023promptbench} and may not always guarantee optimal performance \cite{liu2024adaptive}. 
Despite good results for prompting, finetuning the LLMs still leads to significant performance improvements \cite{mukherjee2024large}.

\section{Evaluation Measures} \label{eval}

A successful style transfer output is one that portrays the correct target style along with preserving the original semantics of the text and maintaining natural language fluency. 

\subsection{Automatic Evaluation} 

Automatic evaluation metrics provide an economic, reproducible, and scalable way to assess the quality of generation results. There are several automated evaluation metrics that have been proposed to measure the effectiveness of TST models \cite{pang2019towards,pang2019daunting,pang2019unsupervised,mir2019evaluating}. They can be divided into three different categories based on the aspect of TST they focus on:


\paragraph{Style Transfer Strength} 
The ability to transfer the text style or the transfer strength of a TST model is measured using Style Transfer Accuracy \cite{hu2017toward,shen2017style,fu2018style,luo2019dual,john2019disentangled}. Mostly, a binary style classifier \cite{DBLP:conf/emnlp/2014} is pre-trained separately to predict the style label of the input sentence and is then used to estimate the style transfer accuracy of the transferred style sentence. This is done by considering the target style as the ground truth.

\paragraph{Content Preservation} 
In order to measure the amount of original content preserved after the style transfer procedure, some automated evaluation metrics from other NLG tasks have been adopted for TST. For instance, the BLEU word-n-gram-overlap metric \cite{papineni2002bleu} is computed similarly as with machine translation.
Match against a target-style sentence can be computed when parallel TST datasets or target-style human references are available. Since most of the TST tasks assume a non-parallel setting and matching references of style transferred sentences are not always a feasible option, evaluation using \textit{source-BLEU (sBLEU)} is adopted. In this method, a transferred sentence is compared to its source. The overlap with the source is considered a proxy for content preservation. Cosine Similarity \cite{rahutomo2012semantic} can also be calculated between the original sentence embeddings and the transfer sentence embeddings \cite{fu2018style}. This methodology follows the idea that the embeddings of the two sentences should be close if most of the semantics are preserved.

\paragraph{Fluency} 
One of the most common goals for all NLG tasks is producing fluent outputs. A common approach to measuring the fluency of a sentence is using a language model \cite{479394}: 
A pre-trained language model is used to compute the perplexity scores of the style-transferred sentences to evaluate the sentences' fluency.

\subsection{Human Evaluation} 

Human evaluation stands out from automatic evaluation due to its flexibility and comprehensiveness. However, this evaluation approach is very challenging since the interpretation of text style can be subjective and vary from individual to individual \cite{pang2019towards,pang2019daunting,mir2019evaluating}. 
In spite of this shortcoming, human evaluations still offer valuable insights into how well the TST algorithms can transfer style and generate sentences that are acceptable according to human standards.

In terms of evaluation types, there is point-wise scoring, wherein humans are asked to provide absolute scores of the model outputs (e.g. on a 1-5 Likert scale), and pairwise comparison, wherein they are asked to judge which of two outputs is better, or by providing a ranking for multiple outputs.

\section{Applications} \label{applications}

TST has a wide range of downstream applications in various NLP fields that include stylized chatbots \cite{mukherjee2023stylized}, stylized writing assistants, automatic text simplification, debiasing online text and even fighting against offensive language. A few very popular examples are discussed below. 

\citet{kim2019comparing} carried out a study that showcased the impact of chatbot's conversational style on users. \citet{DBLP:conf/acl/LiGBSGD16} encoded personas of individuals in contextualized embeddings that helped in capturing the background information and style to maintain consistency in the generated responses. \citet{firdaus2022being} focused on generating polite personalized dialog responses in agreement with the user's profile and consistent with their conversational history.

Another important application of TST is enhancing the human writing experience \cite{can2004change,johnstone2009stance,ashok2013success}. This application aids in people restyling their content to appeal to a variety of audiences, i.e., making a text polite, humorous, professional, or even Shakespearean. 

Another inspiring application of TST is automatically simplifying content for better communication between experts and non-expert individuals in certain knowledge domains, thus lowering language barriers. For example, complicated legal, medical, or technical jargon is transferred into simple terms that a layman can comprehend \cite{cao2020expertise}. 

TST can also offer a means to neutralize subjective attitudes for certain texts where objectivity is strongly needed. For example, in the domains of news, encyclopedia, and textbooks. Such applications can help in reshaping gender roles that are portrayed in writing \cite{clark2018creative}. TST can also help in transforming hateful sentences into non-hateful ones. For instance, \citet{santos2018fighting} propose an extension of a basic encoder-decoder architecture by including a collaborative classifier to deal with abusive language redaction.

\section{Ethical Concerns} \label{ethics}

An essential part of research is to consider the ethical implications of the project through its potential benefits and risks. 

For example, TST has the potential to reduce toxicity, hate speech, sexist and racist language, aggression, harassment, trolling, and cyberbullying \cite{waseem2017understanding}. This task is beneficial for modeling non-offensive text to help reduce toxicity on social media platforms \cite{wired2019twitter,dos2018fighting}. 
It can also be used on social chatbots to make sure there is no bad content in the generated text \cite{roller2020recipes}.
TST is also able to neutralize subjective-toned language, which can be helpful for certain types of publications such as textbooks  \cite{pryzant2020automatically}. 

However, the same technology can also be misused to purposely generate the opposite attribute, i.e., generating hateful, offensive text, that counters any intended social benefit \cite{hovy2016social}. 
Furthermore, as TST is now generally performed using trained language models, these inherit all the potential risks associated with this technology in general, such as reflecting unjust, toxic, or oppressive speech present in the training data \cite{weidinger2022taxonomy}.

The goal of a discussion on ethics is to take into account various concerns like how a system should be built, who it is intended for, and how to assess its societal impact\cite{hovy2016social} \cite{beauchamp2001respect}. Instead of abandoning the whole idea of building such tools, one must explore the concerns and find ways to deal with them \cite{leidner2017ethical}. This should be viewed as an opportunity to increase transparency by surfacing the risks and finding the best ways to its strategy into practice.

\section{Conclusion} \label{conclusion}

The main goal of this work is to offer an introductory overview of Text Style Transfer (TST), highlighting key components such as subtasks, datasets, evaluation methods, and the challenges inherent to TST. Additionally, we discussed the ethical considerations surrounding this area of research. We aim for this overview to serve as a useful guide for those new to the field.

\section*{Acknowledgment}

This research was funded by the European Union  (ERC, NG-NLG,  101039303) and by Charles University projects GAUK 392221 and SVV 260698. 

\bibliography{anthology,custom}
\bibliographystyle{acl_natbib}

\end{document}